\documentclass[10pt, a4paper]{article}
\usepackage{lrec}
\usepackage{graphicx}
\usepackage{tabularx}
\usepackage{soul}
\usepackage{supertabular}

\usepackage{epstopdf}
\usepackage[utf8]{inputenc}
\inputencoding{utf8}
\usepackage{newunicodechar}
\newunicodechar{।}{}

\usepackage[hidelinks]{hyperref}
\usepackage{xstring}
\usepackage{booktabs}

\newcommand* \samethanks[1][\value{footnote}]{\footnotemark[#1]}

\usepackage{breakurl}
\usepackage{multirow}
\usepackage{hyperref}

\title{\textbf{BanFakeNews: A Dataset for Detecting Fake News in Bangla}}

\name{Md Zobaer Hossain$^{\dagger \clubsuit}$\thanks{{ } $\dagger$ First and second authors contributed equally.}, 
Md Ashraful Rahman$^{\dagger \clubsuit}$\samethanks, 
Md Saiful Islam$^\clubsuit$, 
Sudipta Kar$^\spadesuit$
\address{$^\clubsuit$ Shahjalal University of Science and Technology, Sylhet, Bangladesh\\
\texttt{\{zobaer37, ashraful54\}@student.sust.edu, saiful-cse@sust.edu}\\ 
$^\spadesuit$ University of Houston, Texas, USA \\
          \texttt{skar3@uh.edu}\\}}

\abstract{
Observing the damages that can be done by the rapid propagation of fake news in various sectors like politics and finance, automatic identification of fake news using linguistic analysis has drawn the attention of the research community.
However, such methods are largely being developed for English where low resource languages remain out of the focus.
But the risks spawned by fake and manipulative news are not confined by languages.
In this work, we propose an annotated dataset of $\approx50K$ news that can be used for building automated fake news detection systems for a low resource language like Bangla.
Additionally, we provide an analysis of the dataset and develop a benchmark system with state of the art NLP techniques to identify Bangla fake news.
To create this system, we explore traditional linguistic features and neural network based methods.
We expect this dataset will be a valuable resource for building technologies to prevent the spreading of fake news and contribute in research with low resource languages.
The dataset and source code are publicly available at \url{https://github.com/Rowan1697/FakeNews}.

 }

\begin{document}

\maketitleabstract

\section{Introduction}

Articles that can potentially mislead or deceive readers by providing fabricated information are known as fake news. Usually fake news are written and published with the intent to damage the reputation of an agency, entity, or person\footnote{\url{https://en.wikipedia.org/w/index.php?title=Fake_news&oldid=921640983}}.
The popularity of social media (e.g. Facebook), easy access to online advertisement revenue, increased political divergence have been the reasons for the spread of fake news. 
Hostile government actors have also been implicated in generating and propagating fake news, particularly during elections and protests \cite{Fake2} to manipulate people for personal and organizational gains.

Impact of Fake news is creating havoc worldwide.
During the 2016 US election, 25 percent of the Americans visited a fake news website in a six-week period of election which has been hypothesized as one of the issues that influenced the final results \cite{grave2018learning}.
In Bangladesh, the 2012 Ramu incident is an exemplary event where almost 25 thousand people participated in destroying the Buddhist temples on the basis of a Facebook post from a fake account \cite{manik_2012}. 
About  12 Buddhist temples and monasteries and 50 houses were destroyed by the angry mob. Fake news that contains blasphemy can easily repeat these types of incidents where people are very sentimental to their religions.

To tackle fake news there are some dedicated websites like \url{www.politifact.com}, \url{www.factcheck.org}, \url{www.jaachai.com} where they manually update potential fake news stories published in online media with logical and factual explanations behind the news being false.
But these websites are not capable enough as they cannot respond quickly to any fake news event.
Recently computational approaches are also being used to fight against the menace of fake news. 
\newcite{long-etal-2017-fake} tried multi-perspective speaker profiles to detect fake news, where \newcite{yang-etal-2017-satirical} has used linguistic features to detect satirical news.
 Besides \newcite{karadzhov-etal-2017-fully} has proposed a fully automated fact-checking model using external sources to check the claim of news stories. 
 To detect fake news in social media, \newcite{DBLP:journals/corr/abs-1906-05659} has used deep two-path semi-supervised learning.
 However,  these works have been done only on news published in the English. 
 As of now, around 341 millions of people in the world speak Bangla and it is the fifth language in the world in terms of number of speakers \footnote{\url{https://en.wikipedia.org/wiki/List_of_languages_by_number_of_native_speakers}}.
 But to the best of our knowledge, there is no resource or computational approach to tackle the risk of fake news written in Bangla, which can negatively affect this large group of population.

In this paper, we aim at bridging the gap by creating resource for detecting fake news in Bangla.
Our contributions can be summarized as follows:
\begin{itemize}
    \item We publicly release an annotated dataset of $\approx50K$ Bangla news that can be a key resource for building automated fake news detection systems.
    
    \item We develop a benchmark system for classifying fake news written in Bangla by investigating a wide rage of linguistic features. Additionally, we explore the feasibility of different neural architectures and pre-trained Transformers models
    in this problem.
    
    \item We present a thorough analysis of the methods and results and provide a comparison with human performance for detecting fake news in Bangla.

\end{itemize}{}
We expect this work will play a vital role in the development of fake news detection systems. 
In the rest of the paper, we briefly describe our dataset preparation methods, human performance \& observation, and the development of fake news detection systems along with their performance.










\section{Related Works}
Acknowledging the impact of fake news, researchers are trying different methodologies to find a quick and automatic solution to detect fake news in the recent past years. Previous works on satirical news detection mostly use Support Vector Machine (SVM) \cite{rubin-etal-2016-fake,burfoot2009automatic,ahmad2014satire}. Focusing on predictive features an SVM model is proposed by \newcite{rubin-etal-2016-fake} with 360 news articles collected from 2 satirical news sites (The Onion and The Beaverton) and 2 legitimate news sources (The Toronto Star and The New York Times) in 2015 and showed absurdity, grammar and punctuation marks are best for identifying satirical news. Leveraging neural networks, \newcite{yang-etal-2017-satirical} built a 4-level hierarchical network and utilized attention mechanism by using  $\approx16K$ satirical data (collected from 14 satirical news websites) and $\approx160K$ true data and showed paragraph-level features are better than document level features in terms of the stylistic features. Some approaches focused on deception detection and utilised traditional machine learning models such as Naive Bayesian models \cite{oraby2017and} and SVM \cite{ren-zhang-2016-deceptive} to work with linguistic cues. \\
To learn the fake news patterns \newcite{perez-rosas-etal-2018-automatic} also used a linear SVM classifier but to build the supervised learning model they used only linguistic features such as N-grams \cite{Scholkopf:2001:LKS:559923}, Punctuations, LIWC \cite{pennebaker2015development} and conduct their evaluations using five-fold cross-validation. First they collected a dataset of 240 legitimate news from different mainstream news websites in the US then made another dataset containing fake versions of those legitimate news. To generate fake versions of the legitimate news items, they used the crowdsourcing via Amazon Mechanical Turk (AMT). However to find out the underlying features of fake news, neural networks are more useful \cite{shu2017fake} and for textual feature extraction, word embedding technique with deep neural networks are producing quality results. But it is noticeable that neural network based models requires large dataset. Liar \cite{wang2017liar} is a comparatively large dataset containing 12.8K human-labeled short statements collected from POLITIFACT.COM’s API. They used both surface-level linguistic realization and deep neural network architecture. \\
Existing research towards clickbait detection involves hand-crafted linguistic features \cite{chakraborty2016stop,biyani20168} and deep neural networks \cite{gairola2017neural,rony2017diving} with datasets mostly containing clickbaits (headline and article of clickbait news) from different newspapers. Besides \cite{ciampaglia2015computational,vlachos-riedel-2014-fact} have proposed a fact-checking method through knowledge base. And \newcite{karadzhov-etal-2017-fully} has proposed a fully automated fact-checking using external sources where their dataset contains 761 claims from snopes.com, which span various domains including politics, local news, and fun facts. Each of the claims is labeled as factually true (34\%) or as a false rumor (66\%).

Though researches focused on the English language have achieved significant advancement, very few works are available for different low resource languages like Indonesian, Bangla, Hindi. \newcite{pratiwi2017study} created a dataset containing 250 pages of hoax and valid news articles and proposed a hoax detection model using the Naive Bayes classifier for the Indonesian language. In Chinese, a dataset of 50K news is used by \newcite{zhou2015real} for their real-time news certification system on the Chinese micro-blogging website, Sina Weibo\footnote{\url{https://s.weibo.com/}}.

To the best of our knowledge, our work is the first publicly available news dataset in the Bangla  for fake news detection. Throughout our literature review, we found that most of the works introduce a dataset suitable for their research approach and there is some dataset only focused on particular research topics like stance detection\footnote{\url{http://www.fakenewschallenge.org/}}. Since fake news related research for the Bangla are still in its early stage, we design our dataset in a diverse way so that it can be used in multiple lines of research. So we enrich our dataset with clickbaits, satirical, fake and true news with their headline, article, domain and other metadata which is explained briefly in the next section.

\section{A New Dataset for Detecting Fake News Written in Bangla}

To collect a set of authentic news, we select 22 most popular\footnote{We used the Alexa rankings to determine the popularity (\url{www.alexa.com})} and mainstream trusted news portals in Bangladesh.
For collecting fake news we include the following types of news in our dataset.
\begin{itemize}
    \item {Misleading/False Context:} Any news with unreliable information or contains facts that can mislead audiences.
    \item {Clickbait:} News that uses sensitive headlines to grab attention and drive click-throughs to the publisher's website.
    \item {Satire/Parody:} News stories that are intended for entertainment and parody.
 
\end{itemize}
we have collected news from popular websites that publish satire news in Bangla.
While collecting satirical news from these sites we found that most of the sites have the exact same news. 
So after scraping news from these sites, we discarded the duplicates. 
We have collected the misleading or false context type of news from \url{www.jaachai.com} and \url{www.bdfactcheck.com}. 
These two websites provide a logical and informative explanation of fake news that is already published on other sites.
So we have also collected the news that is mentioned on those two sites from the actual publishing sites and make sure that we avoid the duplicates.  
Clickbait is used to grab attention and drive click which eventually increases site visitors and generates revenue for them\footnote{\url{www.webwise.ie/teachers/what-is-fake-news}}.
And we have found that most of the local or less popular sites usually do this. 
To collect clickbaits, we have gone through some of these sites and manually collect potential clickbait news from there.
We call satire, clickbait, and false informative news all commonly as fake news throughout the paper to avoid ambiguity.
We have also collected the following meta-data along with the headlines and content: 
\begin{itemize}
    \item {The domain of the published news site}
    \item {Publication time}
    \item {Category}
\end{itemize}
From our dataset, we got 242 different categories as different publishers categorize the news in their own way.
To generalize it, we took similar categories from different news to map into a single one. 
Finally, we categorize all news from the dataset into the 12 categories (Table \ref{tab:categories}).

\begin{table}[h]
\centering
\begin{tabular}{lll}
\toprule
\textbf{Category} & \textbf{Authentic} & \textbf{Fake}\\
\midrule
Miscellaneous & 2218 & 654 \\
Entertainment & 2636 & 106 \\ 
Lifestyle & 901 & 102\\
National & 18708 & 99 \\
International & 6990 & 91 \\
Politics & 2941 & 90\\
Sports & 6526 & 54\\
Crime & 1072 & 42\\
Education & 1115 & 30\\ 
Technology & 843 & 29\\
Finance & 1224 & 2\\
Editorial & 3504 & 0\\ \bottomrule
\end{tabular}
\caption{Number of news in each category.}
\label{tab:categories}
\end{table}

\begin{figure}[h!]
\centering
\includegraphics[width=1.0\columnwidth]{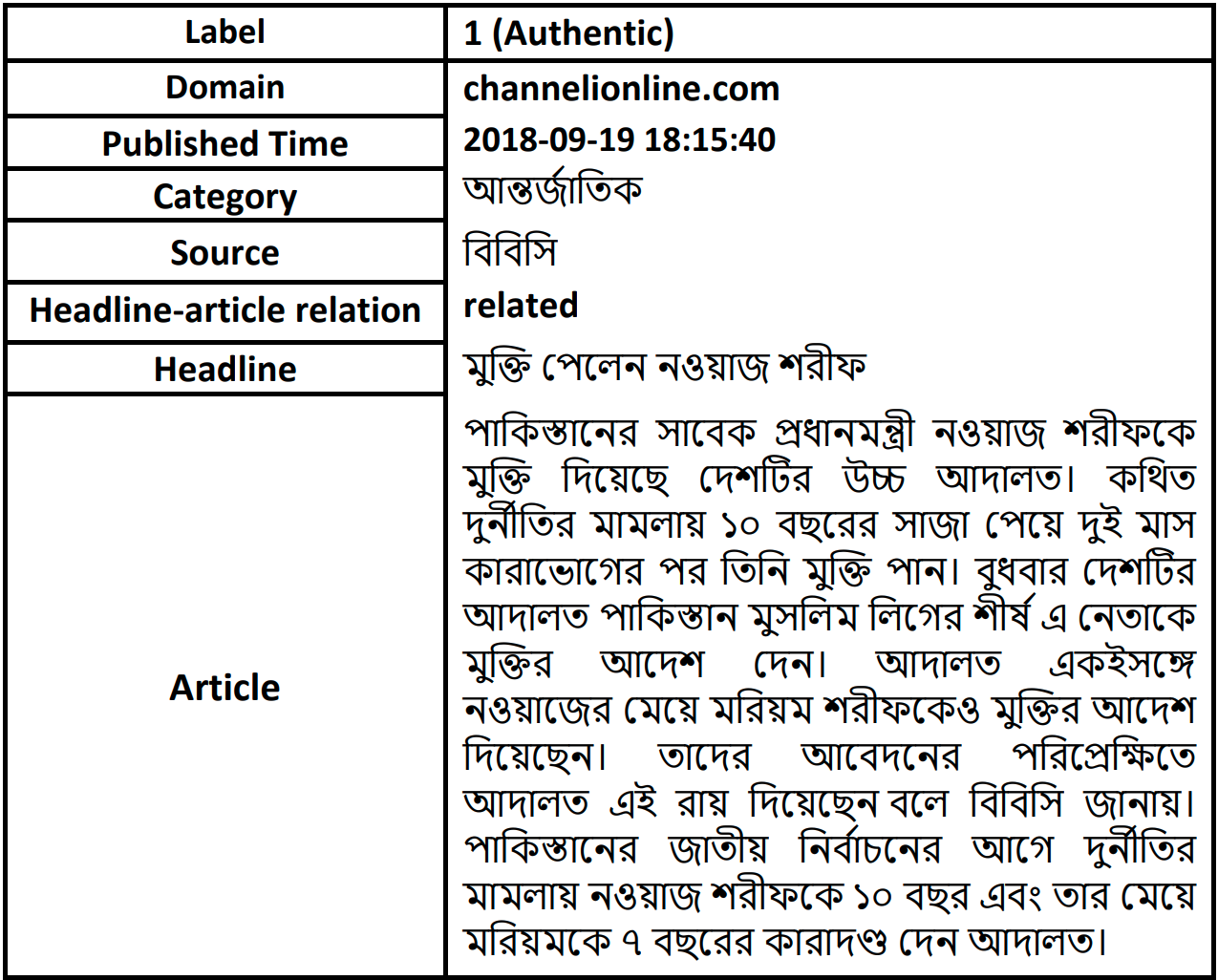}
\caption{Sample Data}
\label{datasetSample}
\end{figure}

Human observation suggests that the source plays a key role in an article’s credibility. 
Note that, by source here we mean one or more person or organization capable of providing verification of the claimed news. 
If there is no such source, then reporters or journalists are taken as the source of news. Besides, \newcite{long-etal-2017-fake} has shown that adding speaker profiles along with document-level features improved the performance of fake news detection.  
So to make our dataset more resourceful, we include the source information as a meta-data for each news. Besides the source, we have also included the headline article relation in meta-data. “Related” and “Unrelated” tags are provided upon checking the relationships of the headline with the article. Since we have to go through each of the news to find out the source and headline-article relation so far we have managed to annotate only $\approx8.5K$ data. All of the members of our data annotator team are undergraduate students of Computer Science and Engineering and Software Engineering department.
Figure \ref{datasetSample} is a sample from our dataset.

\begin{table}[h]
\resizebox{\columnwidth}{!}{
\begin{tabular}{cccccc}
\toprule
          & \textbf{T}     & \textbf{C}       & \textbf{P}     & \textbf{W}      & \textbf{S}\\
          
\midrule
\textbf{Authentic} & 48678 & 1479.14 & 41.20 & 271.16 & 21.15 \\ 
\textbf{Fake}      & 1299  & 1428.19 & 44.13 & 276.36 & 23.62 \\ \bottomrule
\end{tabular}}
\caption{Distribution of mean of characters, punctuations, words, and sentences along with total number of authentic and fake news in dataset. T, C, P, W, S denotes total news count, characters, punctuations, words, sentences mean respectively.}
\label{tab:datasetbynumber}
\end{table}

\section{Human Baseline}
\begin{figure}[t]
\centering
\includegraphics[width=1.0\columnwidth]{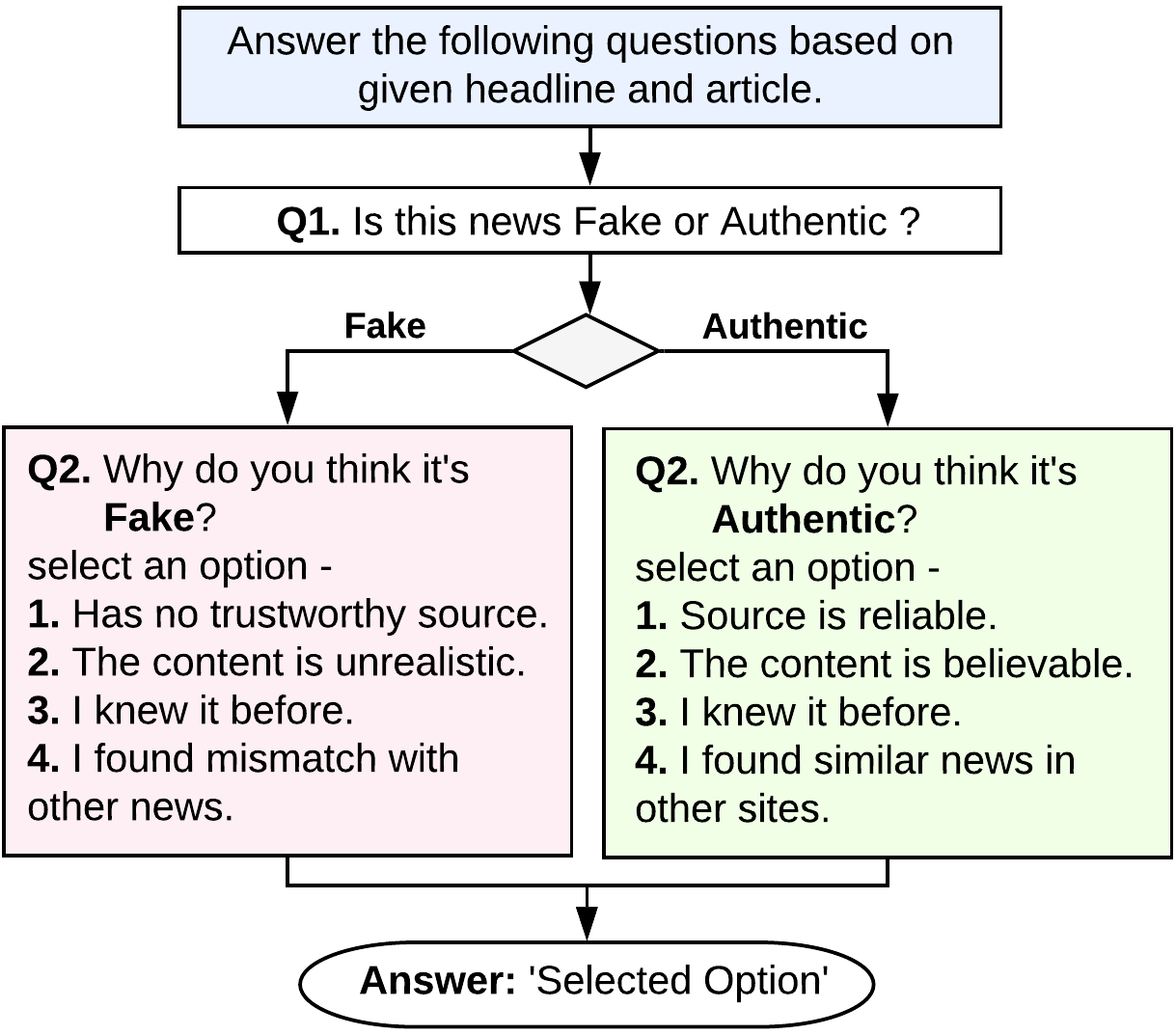}
\caption{Process of human baseline experiment}
\label{humanbaseline}
\end{figure}

Detecting fake news is a tough task. To see how good humans perform, we have conducted an experiment where we took 60 fake news and mixed them randomly with 90 authentic news and gave them to 5 human annotators who are undergraduate students(one Industrial and Production Engineering, one Chemistry and three Computer Science and Engineering students). They were told to read 150 news one by one and answer 2 questions for each news. Note that we only gave them the headlines and the articles. The first question is if the news is fake or authentic. Based on the answer of the first question they were provided another question with four options to select why they think the news is fake or authentic. Here the set of four options are different for fake and authentic. For creating these four options, we first took feedback from 10 people regarding what properties or methods they use to find out if any news is fake or authentic. Then we generalized their feedback into four options. The process of the human baseline experiment is shown in Fig. \ref{humanbaseline} using a flowchart. \\
From the first question, we got an estimate of how accurately humans can detect fake news. The F1-score for the fake class of the five annotators is 58\%, 65\%, 70\%, 68\%, and 63\% respectively. The inter-annotator agreement, measured as Fleiss' Kappa\cite{fleiss1971measuring} is 38.83\% and mean pairwise Cohen's Kappa\cite{cohen1960coefficient} is 39.05\% which indicates that our human annotators have given the same answer on nearly 39\% percent of the questions. The second question helped us to find out what factors are crucial for humans to find the difference between fake and authentic news. If the news is fake, on an average 44\% answer is ‘The content is unrealistic' and 42\% answer is ‘Has no trustworthy source'. When news is true, on an average 62\% answer is ‘The Content is believable' and 21\% answer is ‘Source is reliable'. Here the source is a person or an organization who/that can provide the validation of the claimed news.  From the feedbacks of annotators on overall answers, we found that they choose the option ‘The Content is believable’ because they read or hear similar news on their daily life and the content has nothing to disbelieve it. If the content looks fishy to them then they look for the source of the news. So the experiment shows that source of the news is important. 
After the experiment we had a follow-up interview with the human participants to find out why they make mistakes in detecting fake news. Most significant reasons are as follows:
\begin{itemize}
    \item \textbf{Disguise: }If any fake news is represented just like true news that is the fake news contains enriched information such as strong references, scientific facts and statistics in detail then human mistake it as true news.
    
    \item \textbf{Trending: }People tend to believe any news when they notice a lot of newspapers are reporting/have reported the same news.
    
    \item \textbf{Source: }When a dubious news doesn’t contain any reference (an entity who provided the information) then humans take it as false news.
    
    \item \textbf{Satire: }Humans can detect almost all types of satire news but sometimes some true news also sounds like satire and humans take them as fake news.
\end{itemize}
These analysis indicates that the source plays an influential role in fake news detection. So In our dataset, we manually annotated the source of the news so that in future we can use the source as a momentous feature for detecting fake news.



\section{Methodologies}
In this section, we describe the systems we develop to classify fake news written in Bangla.
Our approaches include traditional linguistic features and multiple neural networks based models.
\subsection{Traditional Linguistic Features}
\begin{itemize}
    \item \textbf{Lexical Features:} Due to the strong performance in various text classification tasks, we extract word n-grams (n=1,2,3) and character n-grams (n=3,4,5) from the news articles. As the weighting scheme we use the term frequency-inverse document frequency (TF-IDF).
    
    \item \textbf{Syntactic Features:} Syntactic structure of texts is often beneficial for understanding particular patterns in documents which eventually help classification problems. So we tag the words of the news articles with their \textit{Parts of Speech} (POS) tags using \cite{Loper02nltk:the}. We use the normalized frequency of different POS tags (\textit{Adjective, Noun, Verb, Demonstrative, Adverb, Pronoun, Conjunction, Particle, Quantifier, Postposition}) as a feature set for each document.
    
    \item \textbf{Semantic Features:} Distributed representations of word and sub-word tokens have shown effectiveness in text classification problems by providing semantic information. So we experiment with pre-trained word embedding, where we represent an article by the mean and the standard deviation of the vector representations of the words in it. We experiment with the Bangla 300 dimensional word vectors pre-trained\footnote{\url{https://fasttext.cc/docs/en/crawl-vectors.html}} with Fasttext \cite{grave2018learning} on \textit{Wikipedia}\footnote{\url{http://wikipedia.org}} and \textit{Common Crawl}\footnote{\url{https://commoncrawl.org/}}, where we have a coverage of 55.21\%.
    Additionally, we experiment with another set of pre-trained 100 dimensional word vectors trained on $\approx20K$ Bangla news by \newcite{ahmad2016bengali} with Word2Vec \cite{Mikolov:2013:DRW:2999792.2999959}, where we have a coverage of 53.95\%. We will call it \textit{News Embedding} throughout the rest of this paper.
    
    \item \textbf{Metadata and Punctuation (MP):} We observed higher presence of some punctuation symbols like `!' in the fake news. So we use the punctuation frequency as features.
    Additionally we use some meta information like the lengths of the headline and the body of news articles as features.
    
 We have found that the publishing sites of fake news are less popular than the sites of true news. So we used the Alexa Ranking\footnote{\url{https://www.alexa.com}} of the sites which are designed to estimate the popularity of websites as a feature. We didn't find the rank of some of the news sites so we annotate these with maximum rank from other sites. And we used the normalized value of ranks as a feature in experiments.

\end{itemize}
\subsection{Neural Network Models}
Neural networks are demonstrating impressive performance in a wide range of text classification and generation tasks.
Given a large amount of training data, such models typically achieve higher accuracy than linguistic feature based methods. 
Hence, we experiment with several neural network models that have been used as benchmark models in different text classification tasks.\\

\textbf{Convolutional Neural Network (CNN):}
Convolutional networks have shown effectiveness in classifying short and long texts in varieties of problems \cite{kim-2014-convolutional,shrestha-etal-2017-convolutional}. 
So we experiment on classifying fake news using a CNN model similar to \cite{kim-2014-convolutional}.
We use kernels having lengths from 1 to 4 and empirically use 256 kernels for each kernel length.
As a pooling layer, we experiment with global max pool and average pool.
We use ReLU \cite{DBLP:journals/corr/abs-1803-08375} as the activation function inside the network.\\

\textbf{Long Short Term Memory:}
Due to the capability of capturing sequential information in an efficient manner, Long Short Term Memory (LSTM) \cite{hochreiter1997long} networks are one of the most widely used models in text classification and generation problems.
Specifically, bidirectional LSTM (Bi-LSTM) have shown impressive performance by capturing sequential information from the both directions in texts.
Moreover, attention mechanism has been seen as a strong pooling technique for classification tasks when used with Bi-LSTM.

In this work, we experiment with a Bi-LSTM model having attention on top which is similar to \cite{zhou-etal-2016-attention}. We use 256 LSTM units. We use two layers of Bi-LSTM in the network.


\subsection{Pre-trained Language Model}
Recently pre-trained language models like OpenAI GPT \cite{radford2018improving}, BERT \cite{DBLP:journals/corr/abs-1810-04805}, ULMFiT \cite{howard2018universal} have made a breakthrough in a wide range of NLP tasks. 
Specifically BERT and its variant models have shown superior performance in the GLUE benchmark for Natural Language Understanding (NLU) \cite{wang2018glue}.
To evaluate the scope of such a language model in our work, we use the multi-lingual BERT model to classify news documents.
We use the pre-trained model weights and implementation publicly distributed by HuggingFace’s Transformers \cite{Wolf2019HuggingFacesTS}.

\section{Experimental Setup}
\textbf{Data Pre-processing:} 
We perform several pre-processing techniques like text normalization and stop words, punctuation removal from the data. Here we use Bangla stop words from Stopwords ISO
 \footnote{\url{https://github.com/stopwords-iso}}.
We observe better validation performance by such pre-processing of the data.


\textbf{Evaluation Metric:} We use Micro-F1 scores to evaluate different methods. As the dataset is imbalanced, we also report the precision (P), recall (R), and F1 score for the minority class (\textit{fake}).

\textbf{Baselines:} We compare our experimental results with a majority baseline and a random baseline. The majority baseline assigns the most frequent class label (\textit{authentic news}) to every article, where the other baseline randomly tags an article as \textit{authentic} or \textit{fake}. We report the mean of precision, recall, and F1-score of 10 random baseline experiments in Table \ref{tab:validation_results}. The Standard Deviation(SD) of precision, recall, and F1-score  in both overall and fake class is less than $10^{-2}$ except the recall of Fake class which is 0.027. 

\textbf{Experiments:}
With the linguistic features, we experiment on training a Linear Support Vector Machine (SVM) \cite{Hearst:1998:SVM:630302.630387}, Random Forest (RF) \cite{RF} and a Logistic Regression (LR)  \cite{McCullagh:1989} model. We split our dataset for training and testing in a 70:30 train-to-test ratio.  
We tune the penalty parameter ($C$) based on the validation results.


For BiLSTM, CNN, and BERT based experiments, the hyper-parameters are Optimizer: Adam Optimizer \cite{DBLP:journals/corr/KingmaB14}, Learning rate: 0.00002, Batch size: 32. Hidden size for BERT model is 756 while in CNN and BiLSTM it is 256. For CNN, we use the kernel lengths of 1 to 4 and zero right paddings in the experiment.  The train and test dataset is kept at a 70:30 ratio. And In training, we use 10\% of the test data as validation data. We use 50 epochs for each experiment and put a checkpoint on F1 score of fake class using validation data. And finally we report the result using our test dataset on the best scoring model from training.

In BERT, For fine tuning our dataset we use the sequence classification model by HuggingFace’s Transformers. And we use the BERT's pre-trained multilingual cased model which is trained on 104 different languages\footnote{\url{https://github.com/google-research/bert}}.


\section{Results and Analysis}
\begin{table}[h]
\centering
\resizebox{\columnwidth}{!}{
\begin{tabular}{l|lll|lll}
    \toprule
        \multicolumn{1}{l}{} & \multicolumn{3}{c}{\textbf{Overall}} & \multicolumn{3}{c}{\textbf{Fake Class}}\\
        \multicolumn{1}{c}{} & \textbf{P} & \textbf{R} & \multicolumn{1}{l}{\textbf{F1}} & \textbf{P} & \textbf{R} & \textbf{F1}\\
        \midrule
        \multicolumn{3}{l}{\textbf{Baselines}}\\\midrule
        Majority  & 0.97 & 1.00 & 0.99 & 0.00 & 0.00 & 0.00\\
        Random  & 0.97  & 0.50 & 0.66 & 0.03 & 0.50 & \textbf{0.05}\\
        \midrule
        \multicolumn{3}{l}{\textbf{Traditional Linguistic Features}}\\\midrule
        Unigram (U) & 0.99 & 0.99 & 0.99 & 0.99 & 0.71 & 0.83\\
        Bigram (B) & 0.98  & 0.99 & 0.99 & 0.97 & 0.42 & 0.59\\
        Trigram (T) & 0.98 & 0.99 & 0.98 & 0.74 & 0.31 & 0.44\\
        U+B+T & 0.99 & 0.99 & 0.99 & 0.98 & 0.68 & 0.80\\
        C3-gram(C3) & 0.99 & 0.99 & 0.99 & 0.98 & 0.82 & 0.89\\
        C4-gram(C4) & 0.99 & 0.99 & 0.99 & 0.99 & 0.78 & 0.87\\
        C5-gram(C5) & 0.99  & 0.99 & 0.99 & 1.00 & 0.74 & 0.85\\
        C3+C4+C5 & 0.99  & 0.99 & 0.99 & 1.00 & 0.77 & 0.87\\
        All Lexical(L) & 0.99 & 0.99 & 0.99 & 1.00 & 0.76 & 0.86\\
        POS tag & 0.97 & 1.00 & 0.98 & 1.00 & 0.00 & 0.01\\
        L+POS & 0.99 & 0.99 & 0.99 & 0.99 & 0.76 & 0.86\\
        Embedding(F) & 0.98 & 0.99 & 0.99 & 0.94 & 0.33 & 0.49\\
        Embedding(N) & 0.98 & 0.99 & 0.99 & 0.84 & 0.32 & 0.46\\
        L+POS+E(F) & 0.99 & 0.99 & 0.99 & 0.99 & 0.77 & 0.87\\
        L+POS+E(N) & 0.99 & 0.99 & 0.99 & 0.98 & 0.79 & 0.88\\
        MP & 0.97 & 0.99 & 0.98 & 0.94 & 0.15 & 0.27\\
        L+POS+E(F)+MP & 0.99 & 0.99 & 0.99 & 0.99 & 0.84 & \textbf{0.91}\\
        L+POS+E(N)+MP & 0.99 & 0.99 & 0.99 & 0.98 & 0.84 & \textbf{0.91}\\
        All Features & 0.99 & 0.99 & 0.99 & 0.98 & 0.84 & \textbf{0.91}\\
        \midrule
        \multicolumn{3}{l}{\textbf{Neural Network Models}}\\\midrule
        CNN & 0.98 & 1.00 & 0.99 & 0.79 & 0.41 & \textbf{0.59}\\
        LSTM & 0.99 & 0.99 & 0.99 & 0.69 & 0.44 & 0.53\\
        BERT & 0.99 & 1.00 & 0.99 & 0.80 & 0.60 & \textbf{0.68}\\ 
        
    \bottomrule
    \end{tabular}}
    \caption{Results of experiments with Traditional Linguistic Features (SVM) and Neural Networks with test set. P and R denote precision and recall, respectively. ‘F' and ‘N' abbreviate ‘Fastext' and ‘News', respectively.}
    \label{tab:validation_results}
\end{table}{}
\begin{figure*}[t]
\centering
\includegraphics[width=12cm]{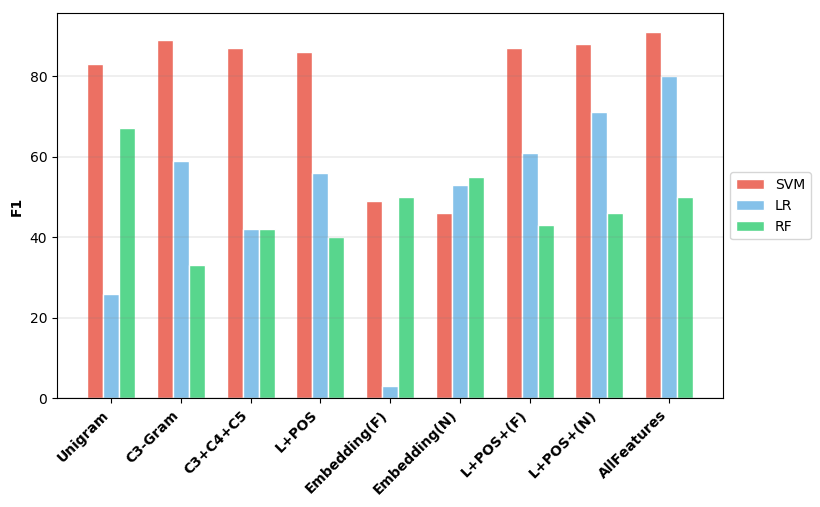}
\caption{Comparison between SVM, LR and RF on linguistic features}
\label{result_barplot}
\end{figure*}
We report our results in Table \ref{tab:validation_results}. Overall performance of every experiment is almost the same. Most of the cases we achieve almost perfect Precision, Recall and F1.  But the results of Precision, Recall, and F1-Score of fake class vary in experiments to experiments.  In our dataset for experiments, the number of authentic news is 37.47 times higher than the number of fake news which could be the reason behind such variance in results of the overall and fake class. To evaluate the performance of different models we will use the precision, recall, F1-Score of the fake class in the rest of the section. 

Fig. \ref{result_barplot} shows that experiment with linguistic features with SVM, RF and LR. Here SVM outperforms LR and RF by a quite margin except result of the news embedding. In the case of news embedding RF scores 55\% of F1-score and here SVM, LR scores 46\%, 53\% of F1-score respectively. For most of the features RF performs better than the LR model. We report the result of SVM on Table \ref{tab:validation_results} since it has performed better than the others.
Table \ref{tab:validation_results} shows that lexical features perform better than other linguistic features, majority \& random baselines, and neural network models as well. It is also observed that the F1-score of fake class in the SVM model decreases while increasing the number of grams in both word and character n-grams. 

The result of POS tag does not improve over the random baselines and the F1-score of this feature indicates that it cannot separate fake news from authentic news. Again F1-Score of MP, Word Embedding(Fasttext), Word Embedding(News) are better than the POS tags but fall behind the lexical features and neural network models. However, these features outperform the majority, and random baselines. We got our best result when incorporating all linguistic features with SVM. It scores 91\% F1-score.

Neural Network models have shown better results in different text classification problems \cite{lai2015recurrent,joulin2016bag}. But in our experiments, we found that F1-Score of fake class in neural networks cannot outperform the linear classifiers. In CNN, experiment with average pooling, global max technique scores 59\% and  54\% F1-Score respectively. Generally an  attention model with LSTM performs better than CNN \cite{yang2016hierarchical,chen-etal-2016-neural-sentiment}. In our case, the F1-Score of BiLSTM with attention cannot improve over the CNN model. The best results of neural networks based experiments came from BERT, whose F1-Score is 68\%. Though neural network models perform better than the majority and random baselines still it falls behind the performance of the SVM model.

To compare the performance of our fake news detection system with humans, We tested the same dataset that we used for the human baseline with our two best models. The model with SVM and character 3-gram scores 89\% F1-score where it scores 90\% with SVM and all linguistic features. Here, the performance of our best models is almost similar to the test set. In our human baseline dataset, there is 90 authentic news along with 60 fake news.  And the model with linguistic features has successfully made correct assumptions on 50 fake news. On the other hand, the character 3-gram model has picked out 49 correct assumptions. Compared to human performance which 64.8\% of F1-score these two models have performed much better. 
These results suggest that linear classifiers could separate the slightest margin of true and authentic news which humans usually overlooked.

\section{Conclusion}
In this paper, we present the first labeled dataset on Bangla Fake news. Here the evaluation of linear classifiers and neural network based models suggest that linear classifiers with traditional linguistic features can perform better than the neural network based models. Another key finding is that character-level features are more significant than the word-level features. In addition to that, it is also found that the use of punctuations in fake news is more frequent than authentic news, and most of the time fake news is found on the least popular sites.

However, since character level features have shown better results so we will incorporate the character level features in neural network models such as \cite{kim2016character,hwang2017character}. From human observation, we found that the source can also play a key role in fake news detection. We will also include this feature in our future experiments. 
Besides we will also continue to expand our dataset. We have manually annotated around 8.5K news. We will continue this process to reach the 50K mark. 
We hope our dataset will provide opportunities for other researchers to use computational approaches to fake news detection.

\section{Acknowledgements}
We would like to thank data annotators and resource team who helped us during human baseline experiment. Also we thank Natural Language Processing Group, Dept. of CSE, SUST for their valuable comments and discussions with us.


\section{Bibliographical References}
\bibliography{lrec2020W.bib}
\bibliographystyle{lrec}

\appendix

\section*{Appendix}
\section{News Sources}

\title{A longtable example}

\bottomcaption{Detailed statistics of the collected news with the domain URL and Alexa ranking(as of 08 March 2020).}
\tablehead
{\toprule
\textbf{Domain}                & \textbf{Rank}   & \textbf{\#News} \\
\midrule}
\begin{center}
\begin{supertabular*}{\columnwidth}{p{4.75cm}lc} 

\multicolumn{3}{l}{\textbf{Authentic News}} \\ 
\url{kalerkantho.com}          & 2040              & 4491            \\
\url{jagonews24.com}          & 1771              & 4426            \\
\url{banglanews24.com}          & 3539              & 4035            \\
\url{banglatribune.com}          & 14505              & 3696            \\
\url{jugantor.com}          & 1422              & 2835            \\
\url{dhakatimes24.com}          & 34756              & 2654            \\
\url{ittefaq.com.bd}          & 6300              & 2589            \\
\url{somoynews.tv}          & 3214              & 2552            \\
\url{dailynayadiganta.com}          & 9678              & 2371            \\
\url{bangla.bdnews24.com}          & 3329              & 2365            \\
\url{prothomalo.com}          & 470              & 2350            \\
\url{bd24live.com}          & 6989              & 2335            \\
\url{risingbd.com}          & 11162              & 2220            \\
\url{dailyjanakantha.com}          & 24403              & 1531            \\
\url{bd-pratidin.com}          & 2141              & 1421            \\
\url{channelionline.com}          & 8878              & 1401            \\
\url{samakal.com}          & 8698              & 1372            \\
\url{independent24.com}          & 216950              & 1220            \\
\url{rtnn.net}          & 1921350              & 1149            \\
\url{bangla.thereport24.com}          & 219278              & 859            \\
\url{mzamin.com}          & 8715              & 785            \\
\url{bhorerkagoj.net}          & 60018              & 21            \\
\midrule
\multicolumn{3}{l}{\textbf{Satire News}}                               \\
\url{channeldhaka.news}          & 249033              & 436            \\
\url{earki.com}          & 2226986              & 291            \\
\url{motikontho.wordpress.com}          & 7291230              & 195            \\
\url{bengalbeats.com}          & 707465              & 192            \\
\url{sarcasmnews.fun}          &  N\textbackslash{}A              & 14            \\
\url{ctnews7}          & 7484275.0              & 4            \\
\url{Prothombarta.news}          & 193562              & 1            \\
\url{The Report24.com}          & 219278             & 1            \\
\url{aparadhchokh24bd.com}          & 4242479              & 1            \\
\url{shadhinbangla24}          & 1115208              & 1            \\

\midrule
\multicolumn{3}{l}{\textbf{Clickbait}}
                     \\
\url{bengaliviralnews.com}          & 1884782              & 48            \\
\url{gonews24.com}          & 83988              & 10            \\
\url{lastnewsbd.com}          & 77165              & 3            \\
\url{bangladeshonline24.com}          & 3023578              & 2            \\
\url{news.zoombangla.com}          & 5770              & 2            \\
\url{prothombarta.news}          & 193562              & 2            \\
\url{prothombhor.net}          & 2831785             & 2            \\
\url{somoyerkonthosor.com}          & 72800              & 2            \\
\url{agooannews.com}          &  N\textbackslash{}A              & 1            \\
\url{aparadhchokh24bd.com}          & 4242479              & 1            \\
\url{banglanews24.com}          & 3527              & 1            \\
\url{bdjournal365.com}          & 2395727              & 1            \\
\url{bdsangbad.com}          & 5272008              & 1            \\
\url{bdtype.com}          & 309252              & 1            \\
\url{bn.mtnews24.com}          & 63562              & 1            \\
\url{daily-bangladesh.com}          & 6961              & 1            \\
\url{dkpapers.com}          & 216868              & 1            \\
\url{sangbadprotidin24.com}          & 1489949              & 1            \\
\url{sonalinews.com}          & 37689             & 1            \\
\midrule
\multicolumn{3}{l}{\textbf{Fake}}
\\
\url{banglainsider.com}          & 23339              & 3            \\
\url{bd-pratidin.com}          & 2141              & 3            \\
\url{bengaliviralnews.com}          & 1884782              & 3            \\
\url{notunshokal.com}          & 3012897              & 3            \\
\url{alokitobangladesh.com}          & 84832              & 2            \\
\url{bangla.dhakatribune.com}          & 15159              & 2            \\
\url{banglanews24.com}          & 3527              & 2            \\
\url{dailyinqilab.com}          & 9970              & 2            \\
\url{dailysangram.com}          & 84689              & 2            \\
\url{ittefaq.com.bd}          & 6285              & 2            \\
\url{jugantor.com}          & 1420              & 2            \\
\url{kalerkantho.com}          & 2031              & 2            \\
\url{shadhinbangla24.com.bd}          & 1268056              & 2            \\
\url{somewhereinblog.net}          & 50584              & 2            \\
\url{sylhettoday24.news}          & 94555              & 2            \\
\url{bangla.bdnews24.com}          & 3332              & 1            \\
\url{bangla24.com.bd}          & 6652978              & 1            \\
\url{bangladeshbani24.com}          & N\textbackslash{}A               & 1            \\
\url{banglatribune.com}          & 14517              & 1            \\
\url{bd-journal.com}          & 8334              & 1            \\
\url{bd24live.com}          & 6989              & 1            \\
\url{bd24report.com}          & 23915              & 1            \\
\url{bdhotnews.com}          & 8415177              & 1            \\
\url{bengali.oneindia.com}          & 1150              & 1            \\
\url{bn.banglafact.com}          & 2911410              & 1            \\
\url{bn.bdcrictime.com}          & 57569              & 1            \\
\url{bn.mtnews24.com}          & 63562              & 1            \\
\url{channelionline.com}          & 336950              & 1            \\
\url{city24news.com}          & 147311              & 1            \\
\url{coxsbazarnews.com}          & 130112              & 1            \\
\url{dailyamadernandail.com}          & 3304326              & 1            \\
\url{dailyjanakantha.com}          & 24403              & 1            \\
\url{dailynayadiganta.com}          & 9689              & 1            \\
\url{dailysatkhira.com}          & 738489              & 1            \\
\url{deshebideshe.com}          & 28828              & 1            \\
\url{dhakajournals.com}          & 3625884              & 1            \\
\url{dhakatimes24.com}          & 35,273              & 1            \\
\url{ekushey-tv.com}          & 21121              & 1            \\
\url{famousnews24.com}          & 469905              & 1            \\
\url{gonews24.com}          & 83988              & 1            \\
\url{jagonews24.com}          & 1771              & 1            \\
\url{keuamaremairala.com}          & 10045257              & 1            \\
\url{mujibsenanews.com}          & 2420840              & 1            \\
\url{nirapadnews.com}          & 158811              & 1            \\
\url{ppbd.news}          & 29471              & 1        \\    
\url{priyo.com}          & 38831              & 1            \\
\url{ekusherbangladesh.com.bd}          & 124277              & 1            \\
\url{probashkotha.com}          & 654309              & 1            \\
\url{prothombhor.net}          & 2831785              & 1            \\
\url{protissobi.com}          & 4943617              & 1            \\
\url{rtvonline.com}          & 12281              & 1            \\
\url{sharebusiness24.com}          & 566064              & 1            \\
\url{sharenews24.com}          & 98406              & 1            \\
\url{shawdeshbhumi.com}          & N\textbackslash{}A               & 1            \\
\url{snpsports24.com}          & 1113116              & 1            \\
\url{somoyerkonthosor.com}          & 72800              & 1            \\
\url{sylhetprotidin24.com}          & 3086838              & 1            \\
\url{timesbangla.in}          & 2675956              & 1            \\
\url{awarenessbulletin.blog-spot.com}          & N\textbackslash{}A               & 1            \\
\url{bdexclusivenews.blog-spot.com}          & N\textbackslash{}A               & 1            \\
\url{bangla.24livenews-paper.com}          & 6725              & 1            \\
\url{ourevergreen-bangladesh.com}          & 436458              & 1            \\
\end{supertabular*}
\end{center}




\end{document}